\documentclass[
]{ceurart}

\usepackage{graphicx}

\usepackage{listings}
\usepackage{svg}

\begin{document}

\copyrightyear{2025}
\copyrightclause{Copyright for this paper by its authors.
  Use permitted under Creative Commons License Attribution 4.0
  International (CC BY 4.0).}

\conference{CLEF 2025 Working Notes, 9 -- 12 September 2025, Madrid, Spain}
\title{AI Wizards at CheckThat! 2025: Enhancing Transformer-Based Embeddings with Sentiment for Subjectivity Detection in News Articles}

\title[mode=sub]{Notebook for the CheckThat! Lab at CLEF 2025}


\author[1]{Matteo Fasulo}[%
orcid=0000-0002-7019-3157,
email=matteo.fasulo@studio.unibo.it,
url=https://github.com/MatteoFasulo,
]
\cormark[1]
\fnmark[1]
\address[1]{Department of Computer Science and Engineering (DISI) - University of Bologna}

\author[1]{Luca Babboni}[%
orcid=0009-0001-5260-7467,
email=luca.babboni2@studio.unibo.it,
url=https://github.com/ElektroDuck,
]
\fnmark[1]

\author[1]{Luca Tedeschini}[%
orcid=0009-0006-0375-829X,
email=luca.tedeschini3@studio.unibo.it,
url=https://github.com/LucaTedeschini,
]
\fnmark[1]

\cortext[1]{Corresponding author.}
\fntext[1]{These authors contributed equally.}

\begin{abstract}
This paper presents \textbf{AI Wizards}' participation in the CLEF 2025 CheckThat! Lab Task 1: Subjectivity Detection in News Articles, classifying sentences as subjective/objective in monolingual, multilingual, and zero-shot settings.
Training/development datasets were provided for Arabic, German, English, Italian, and Bulgarian; final evaluation included additional unseen languages (e.g., Greek, Romanian, Polish, Ukrainian) to assess generalization.
Our primary strategy enhanced transformer-based classifiers by integrating sentiment scores, derived from an auxiliary model, with sentence representations, aiming to improve upon standard fine-tuning.
We explored this sentiment-augmented architecture with mDeBERTaV3-base, ModernBERT-base (English), and Llama3.2-1B.
To address class imbalance, prevalent across languages, we employed decision threshold calibration optimized on the development set. Our experiments show sentiment feature integration significantly boosts performance, especially subjective F1 score. This framework led to high rankings, notably 1st for Greek (Macro F1~$=~0.51$).
\end{abstract}


\maketitle

\section{Introduction}

Our work addresses subjectivity detection as defined in Task 1~\cite{clef-checkthat:2025:task1} of the CLEF 2025 CheckThat! Lab. An overview of the CheckThat! Lab and its constituent tasks can be found in~\cite{CheckThat:ECIR2025, clef-checkthat:2025-lncs}. Specifically, Task 1 challenges systems to classify sentences from news articles as subjective (SUBJ) or objective (OBJ). This capability is vital for efforts to combat misinformation and improve fact-checking, as the ability to separate opinion from factual claims is essential, particularly given the rapid growth of multilingual online content. A key difficulty in this task lies in its sentence-level granularity, requiring classification without the wider context of the full article. Historically, subjectivity detection approaches, from lexicon-based methods to classical machine learning~\cite{kamal2013subjectivityclassificationusingmachine}, faced limitations with linguistic variety and subtlety. While contemporary transformer-based models offer substantial advancements, their deployment in diverse multilingual and resource-constrained environments continues to pose challenges.

\noindent
This paper presents our system, which fine-tunes transformer-based models by strategically fusing external sentiment information. We augment sentence representations with sentiment scores from an auxiliary model before classification. This sentiment-enhanced strategy is evaluated on:
\begin{itemize}
    \item mDeBERTaV3-base~\cite{he2021deberta, he2021debertav3}, ModernBERT-base~\cite{modernbert}, and Llama3.2-1B~\cite{grattafiori2024llama3herdmodels} fine-tuned with and without our sentiment feature fusion for multilingual subjectivity detection.
    \item The systematic integration of sentiment scores as a key feature engineering step, demonstrating its impact on improving subjective content classification.
    \item The application of decision threshold calibration to mitigate class imbalance inherent in the provided datasets, further refining performance.
\end{itemize}
We evaluate our system across monolingual, multilingual, and zero-shot settings, focusing on improving the F1 score for the subjective class. Our work aims to provide insights into effective strategies for multilingual subjectivity detection, highlighting benefits of integrating sentiment features and careful handling of imbalanced data within a transformer-based framework.

\section{Related Work}

Subjectivity detection, often used as a preprocessing step to sentiment analysis~\cite{wilson2004just}, aims to filter out objective content and retain subjective sentences, which are then analyzed for polarity. While the two tasks are closely intertwined and can function complementarily~\cite{naveed2022subjectivity}, this pipeline-based approach has been common in early works. Subjectivity detection initially relied on lexical resources (e.g., SentiWordNet~\cite{baccianella-etal-2010-sentiwordnet}) and rule-based systems. While interpretable, these methods lacked adaptability to diverse linguistic expressions and contexts. This limitation was partially addressed by machine learning techniques leveraging engineered features (e.g., n-grams, POS tags), which, however, still faced generalization issues. 

\noindent
The advent of deep learning, particularly transformer-based models like BERT~\cite{devlin2019bertpretrainingdeepbidirectional}, has significantly advanced NLP tasks, including both subjectivity detection and sentiment classification. These models learn rich contextual representations from large unlabeled corpora, enabling superior performance when fine-tuned. Our work aligns with this literature by combining both perspectives: since our goal is to identify subjective sentences, we leverage sentiment analysis signals to reinforce subjectivity predictions—an approach supported by prior findings that highlight the strong interdependence between subjectivity and sentiment~\cite{naveed2022subjectivity}. Additionally, previous CLEF CheckThat! Labs have also demonstrated the effectiveness of transformer architectures for related subtasks, such as identifying subjective claims in news articles~\cite{5519ef103bb441eda94836528e123131}.

\noindent
Multilingual subjectivity detection introduces further complexities. While models like mBERT~\cite{devlin2019bertpretrainingdeepbidirectional} or XLM-R~\cite{conneau2020unsupervisedcrosslingualrepresentationlearning} provide strong cross-lingual transfer baselines, their performance varies across language pairs and task specificities. mDeBERTaV3, with its disentangled attention mechanism~\cite{he2021deberta, he2021debertav3}, has shown strong performance on NLU benchmarks, making it suitable here. More recent models like ModernBERT~\cite{modernbert} aim for comparable performance with improved efficiency, often focusing on English.

\noindent
Augmenting text representations with auxiliary information, like sentiment or emotion, for improved classification is an active research area. Similar to the use of emotions in sexism detection~\cite{muti2023enriching}, we hypothesize that explicit sentiment signals can help disambiguate subjective statements.
Addressing class imbalance is another crucial aspect, especially as one class is often more prevalent in real-world datasets. Techniques range from data-level resampling to algorithmic approaches like cost-sensitive learning or threshold adjustment~\cite{abdelhamid2024balancingscalescomprehensivestudy}. Our decision threshold calibration aligns with findings that post-hoc adjustments can effectively improve performance on imbalanced datasets without altering the training process.

\section{Dataset}

The data for this task is provided by the CLEF 2025 CheckThat! Lab Task 1 organizers.\footnote{\href{https://gitlab.com/checkthat_lab/clef2025-checkthat-lab/-/tree/main/task1/data}{CLEF 2025 CheckThat! Lab Task 1 Data}} The dataset consists of sentences extracted from news articles across five languages: Arabic (AR), Bulgarian (BG), English (EN), German (DE), and Italian (IT). Each sentence is labeled as either subjective (SUBJ) or objective (OBJ). The annotation guidelines, as described in~\cite{antici-etal-2024-corpus}, define subjective sentences as "those expressing personal opinions, sarcasm, exhortations, discriminatory language, or rhetorical figures conveying an opinion. Objective sentences include factual statements, reported third-party opinions, open-ended comments, and factual conclusions".
For each language, the data is split into training, development (dev), and development-test (dev-test) sets. An analysis of the label distribution (Table \ref{tab:dataset_statistics} in Section \ref{sec:experiments_results}) reveals a notable class imbalance across all languages, with the objective class being more frequent. Italian and Arabic exhibit the most pronounced imbalance. This characteristic significantly influences model training and evaluation, necessitating strategies to mitigate its impact. 

\begin{table*}[h!]
  \caption{Distribution of objective (OBJ) and subjective (SUBJ) labels across different languages and dataset splits. The table presents statistics for the training, development (Dev), and development-test (Dev-Test) sets.}
  \label{tab:dataset_statistics}
  \begin{tabular}{l|rrrrrr}
    \toprule
    \textbf{Language} & \multicolumn{2}{c}{\textbf{Training}} & \multicolumn{2}{c}{\textbf{Dev}} & \multicolumn{2}{c}{\textbf{Dev-Test}}\\
                      & \textbf{OBJ} & \textbf{SUBJ} & \textbf{OBJ} & \textbf{SUBJ} & \textbf{OBJ} & \textbf{SUBJ}\\ 
    \midrule
    Arabic           & 1,391        & 1,055         & 266          & 201           & 425          & 323\\
    Bulgarian        & 406          & 323           & 175          & 139           & 143          & 107\\
    English          & 532          & 298           & 240          & 222           & 362          & 122\\
    German           & 492          & 308           & 317          & 174           & 226          & 111\\
    Italian          & 1,231        & 382           & 490          & 177           & 377          & 136\\
    \bottomrule
  \end{tabular}%
\end{table*}

\section{Methodology}

\begin{figure}[h!]
    \centerline{\includegraphics[width=0.8\linewidth]{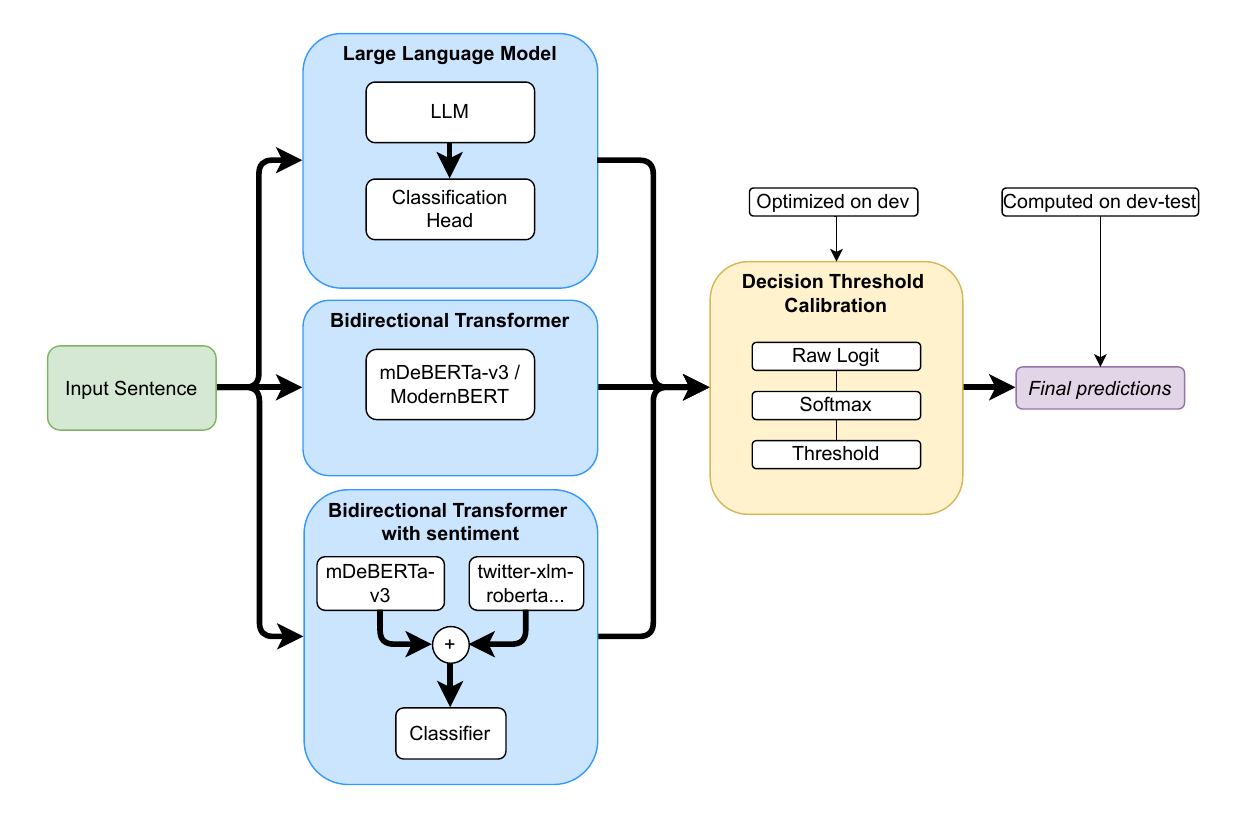}}
    \caption{Model architecture with the three developed pipelines.}
    \label{fig:architecture}
\end{figure}

Our methodology fine-tunes pre-trained transformer models for binary subjectivity classification. A core architectural element is fusing sentiment features with sentence representations before the classification layer. We explore this sentiment-enhanced fine-tuning with several transformer architectures (detailed in Section~\ref{ssec:model_architectures}). To address class imbalance, we implement decision threshold calibration (Section~\ref{ssec:threshold_calib}). An alternative, Focal Loss, is discussed in Appendix~\ref{sec:focal-loss}. The general pipeline is illustrated in Figure~\ref{fig:architecture}. All fine-tuning used a Kaggle environment with a single NVIDIA Tesla P100 GPU (16GB VRAM).

\subsection{Model Architectures}
\label{ssec:model_architectures}

We experiment with three main types of transformer-based models:
\begin{itemize}
    \item mDeBERTaV3-base: A powerful multilingual model chosen for its strong cross-lingual generalization capabilities, essential for handling the diverse languages in the task.
    \item ModernBERT-base: A more recent English-centric model designed for efficiency and performance. We evaluate this primarily for the English monolingual task.
    \item Llama3.2-1B: A smaller-scale Large Language Model. We adapt this by adding a classification head and fine-tuning it, primarily for English, to compare its capabilities against BERT-like architectures on this specific task. Due to resource constraints on the environment, this model was fine-tuned using 8-bit quantization with LoRA as to fit inside a single P100 GPU.
\end{itemize}

\noindent
For all models, a standard classification head (a simple feed-forward neural network) is added on top of the [CLS] token representation (or the equivalent final hidden state for Llama).

\subsection{Sentiment Augmentation}

To provide the models with explicit signals about the affective content of a sentence, which we hypothesize correlates with subjectivity, we incorporate sentiment scores as additional features.
\begin{itemize}
    \item Sentiment Prediction: For each input sentence, we first predict its sentiment using an external pre-trained multilingual sentiment analysis model, twitter-xlm-roberta-base-sentiment~\cite{barbieri-etal-2022-xlm}. This model outputs a three-dimensional vector representing probabilities for positive, neutral, and negative sentiment. It was selected primarily for its robust multilingual capabilities and its widespread adoption in sentiment analysis tasks, despite its training domain (Twitter data) being different from our context of news articles.
    \item Feature Concatenation: These three sentiment scores are then concatenated with the [CLS] token embedding (the output of the base transformer model) before being passed to the final classification layer. This effectively expands the input dimensionality of the classifier to include both the learned textual representation and the explicit sentiment signal. This approach was primarily applied with the mDeBERTaV3-base model.
\end{itemize}

\subsection{Data Preprocessing and Tokenization}
\label{ssec:preprocessing}

Sentences are tokenized using the specific tokenizer associated with each pre-trained model (mDeBERTa, ModernBERT, Llama). We apply padding and truncation to a maximum sequence length of 256 tokens, which covers the majority of sentence lengths in the datasets (more than 75\% of sentences lenght). Recognizing potential performance disparities across languages when using multilingual models, and with a view to addressing specific complexities that might arise with languages like Arabic (which, as we will discuss, presented challenges), we explored an additional strategy for the Arabic experiments. This involved translating the Arabic data into English using the Helsinki-NLP/opus-mt-ar-en model ~\cite{tiedemann2023democratizing, tiedemann-thottingal-2020-opus} prior to fine-tuning. The aim was to assess if this could mitigate some of the language-specific difficulties, though this particular avenue did not ultimately lead to improved performance in our final configuration while giving slightly worse results. We attribute this outcome to several potential factors: (1) inaccuracies and loss of fidelity introduced by the machine translation process; (2) the inherent difficulty in preserving subtle, culturally-specific linguistic nuances crucial for subjectivity detection when translating from Arabic to English; and (3) a resultant mismatch in sentiment representation, as the sentiment features for this experimental branch would have been derived from the translated English text, potentially not reflecting the original Arabic sentiment accurately.

\subsection{Training and Decision Threshold Calibration}
\label{ssec:threshold_calib}

Models are fine-tuned using the AdamW optimizer with a linear learning rate scheduler and warmup, employing Cross-Entropy Loss with class weights to initially mitigate class imbalance. Batch size was 16, learning rate $1 \times 10^{-5}$, for 6 epochs. The best checkpoint is selected based on development set performance.

\noindent
Addressing the challenge of substantial class imbalance, especially concerning the subjective class, we employed a post-hoc decision threshold optimization strategy. Initially, the model is trained on the training set using cross-entropy loss. We then select the best-performing checkpoint based on development set metrics. For this checkpoint, an optimal decision threshold is determined by conducting a grid search over values ranging from $0.1$ to $0.9$ ($0.01$ increment), aiming to maximize the macro F1 score on the development set. Finally, this optimized threshold is applied to the model's softmax outputs for classification on the test set. This procedure allows for fine-tuning the decision boundary to the dataset's class distribution while ensuring proper methodological separation between training, development, and testing phases, thereby guarding against overfitting to the test set.

\section{Experiments and Results}
\label{sec:experiments_results}

We conducted experiments for the monolingual, multilingual, and zero-shot subjectivity detection subtasks defined by CLEF 2025 CheckThat! Lab Task 1~\cite{clef-checkthat:2025:task1}. Evaluation primarily focuses on macro-average F1 and SUBJ F1 scores, given the latter's importance amidst class imbalance. All reported dev-test results utilize the decision threshold calibration from Section~\ref{ssec:threshold_calib}.

\subsection{Monolingual Task}

In the monolingual setting, models were trained and evaluated on each language independently (Table \ref{tab:monolingual}). mDeBERTaV3-base generally performed well, particularly for German and Italian. Adding sentiment features (mDeBERTa-V3-sentiment) consistently improved SUBJ F1 scores across most languages, with notable gains for English (0.4046 to 0.5279) and Italian (0.6291 to 0.6804), suggesting sentiment information provides valuable cues for subjective content. ModernBERT (English only) was competitive, slightly outperforming baseline mDeBERTaV3-base on English SUBJ F1. Llama3.2-1B, even with LoRA, did not match BERT-like architectures for English. Pre-translating Arabic data into English (Section~\ref{ssec:preprocessing}) did not improve results and was not pursued for final models.

\begin{table}[h!]
\caption{Model performance on monolingual setting across languages. The table reports Macro F1 and Subjective F1 scores using the decision threshold calibration procedure.}
\label{tab:monolingual}
\begin{tabular}{lcc}
  \toprule
  \textbf{Language} & \textbf{Macro F1} & \textbf{SUBJ F1} \\
  \midrule
  \multicolumn{3}{l}{\textbf{mDeBERTa-V3}} \\ 
  Arabic     & 0.5805 & 0.5598 \\ 
  Bulgarian  & 0.7555 & 0.7222 \\ 
  English    & 0.6650 & 0.4843 \\ 
  German     & 0.8218 & 0.7652 \\ 
  Italian    & 0.7654 & 0.6291 \\
  \midrule
  \multicolumn{3}{l}{\textbf{mDeBERTa-V3 + Sentiment}} \\
  Arabic     & 0.5735 & 0.5741	  \\
  Bulgarian  & 0.7718 & 0.7407	  \\
  English    & 0.7036 & 0.5279 \\ 
  German     & 0.8291 & 0.7759	 \\ 
  Italian    & 0.7769 & 0.6804 \\
  \midrule
  \multicolumn{3}{l}{\textbf{ModernBERT}} \\ 
  English    & 0.6922 & 0.5612 \\
  \midrule
  \multicolumn{3}{l}{\textbf{Llama3.2-1B}} \\ 
  English    & 0.6375 & 0.4046 \\
  \bottomrule
  \end{tabular}%
\end{table}

\paragraph{Impact of Threshold Calibration}

Table \ref{tab:thresholding} demonstrates the impact of the decision threshold calibration. For languages with significant class imbalance like Arabic and Italian, calibration leads to substantial improvements in both Macro F1 and SUBJ F1 scores. For more balanced languages (e.g., Bulgarian, German), the gains are marginal or, in some cases like English for mDeBERTa-V3 baseline, standard thresholding performed slightly better by one metric, indicating the complexity of interaction between model, data distribution, and thresholding. Overall, however, calibration proved beneficial, especially for the target SUBJ class in imbalanced scenarios.

\begin{table}[h!]
  \caption{Comparison of model performance using decision threshold calibration procedure across different languages. The table reports Macro F1 and Subjective F1 scores. Here we refer to the base models not using sentiment values.}
  \label{tab:thresholding}
  \begin{tabular}{lcccc}
  \toprule
  \textbf{Language} & \multicolumn{2}{c}{\textbf{Threshold}} & \multicolumn{2}{c}{\textbf{No Threshold}} \\
                    & \textbf{Macro F1} & \textbf{SUBJ F1} & \textbf{Macro F1} & \textbf{SUBJ F1} \\
  \midrule
  Arabic     & 0.5805 & 0.5598 & 0.5538 & 0.4184 \\ 
  Bulgarian  & 0.7555 & 0.7222 & 0.7491 & 0.6970 \\ 
  English    & 0.6650 & 0.4843 & 0.6610 & 0.4775 \\ 
  German     & 0.8218 & 0.7652 & 0.8217 & 0.7699 \\ 
  Italian    & 0.7654 & 0.6291 & 0.7048 & 0.6237 \\
  \bottomrule
  \end{tabular}%
\end{table}

\subsection{Multilingual and Zero-Shot Tasks}

For the multilingual task, mDeBERTaV3-base was fine-tuned on a combined dataset of all languages. The model achieved a Macro F1 of 0.6942 and a SUBJ F1 of 0.6114 (Table \ref{tab:multilingual}). When Arabic was excluded from the training and evaluation (given its consistently challenging nature), performance on the remaining languages improved to a Macro F1 of 0.7817 and SUBJ F1 of 0.6887. Adding sentiment features in the multilingual setting (mDeBERTa-V3 + Sentiment) showed mixed results when all languages were included but provided the best performance when Arabic was excluded (Macro F1 0.7962, SUBJ F1 0.7114).

\noindent
In the zero-shot setting, where models were trained on a subset of languages and tested on unseen ones, performance varied depending on the specific language combinations. Generally, models performed better when the training set included linguistically diverse languages or those with larger datasets. The challenges observed with Arabic in monolingual and multilingual settings persisted in zero-shot scenarios, often leading to lower performance when Arabic was a target unseen language. Detailed zero-shot results (e.g., Table \ref{tab:zeroshot}) indicate that achieving robust generalization to entirely unseen languages remains a significant challenge, though sentiment augmentation sometimes provided benefits.

\begin{table}[h!]
\caption{Evaluation results of mDeBERTa-V3 on multilingual data.}
\label{tab:multilingual}
  \begin{tabular}{lcc}
  \toprule
  \textbf{Language} & \textbf{Macro F1} & \textbf{SUBJ F1} \\
  \midrule
  \multicolumn{3}{l}{\textbf{mDeBERTa-V3}} \\ 
  Multilingual   & 0.6942 & 0.6114 \\ 
  Excluding Arabic & 0.7817 & 0.6887 \\
  \midrule
  \multicolumn{3}{l}{\textbf{mDeBERTa-V3 + Sentiment}} \\
  Multilingual   & 0.6798 & 0.5332 \\
  Excluding Arabic & 0.7962 & 0.7114 \\
  \bottomrule
  \end{tabular}%
\end{table}

\begin{table}[h!]
\caption{Zero-shot performance. Models were trained on the 'Training Languages' and tested on the remaining languages from the initial set of five (Arabic, Bulgarian, English, German, Italian) not included in the training set for that row.}
\label{tab:zeroshot}
    \begin{tabular}{llcc}
    \toprule
    \textbf{Training languages}      & \textbf{Model}          & \textbf{Macro F1} & \textbf{SUBJ F1}  \\
    \midrule
Ar, Bg, Ge       & mDeBERTaV3              & 0.7395   & 0.6066   \\
Ar, Bg, Ge       & mDeBERTaV3 + Sentiment  & 0.7461   & 0.6134   \\
En, It                & mDeBERTaV3              & 0.6147   & 0.5166   \\
En, It                & mDeBERTaV3 + Sentiment  & 0.6121	   & 0.5087   \\
  \bottomrule
  \end{tabular}
\end{table}
\newpage
\subsection{Analysis of Sentiment Augmentation}
The positive impact of sentiment augmentation, especially for English and Italian SUBJ F1 scores, warrants further investigation. As detailed in our discussion, we observed that sentences correctly classified as subjective by the sentiment-enhanced model (but misclassified by the baseline) often exhibited stronger negative sentiment scores (Table \ref{tab:sentiment_right} and \ref{tab:sentiment_wrong}). This suggests the model learns to associate pronounced sentiment (particularly negative, in the context of news critique or opinion) with subjectivity. The distribution of sentiment scores across the dataset further indicates a tendency for subjective sentences to carry more polarized sentiment.

\begin{table}[h!]
\caption{Mean and standard deviation of sentiment values when the sentiment model correctly identifies sentences, but the other model fails.}
\label{tab:sentiment_right}
\begin{tabular}{llcccccc}
\toprule
\textbf{Label} & & \multicolumn{3}{c}{\textbf{Mean}} & \multicolumn{3}{c}{\textbf{Std}} \\
\cmidrule(lr){3-5} \cmidrule(lr){6-8}
& & \textbf{Positive} & \textbf{Neutral} & \textbf{Negative} & \textbf{Positive} & \textbf{Neutral} & \textbf{Negative} \\
\midrule
\textbf{OBJ} & & 0.32 & 0.31 & 0.36 & 0.20 & 0.31 & 0.32 \\
\textbf{SUBJ} & & 0.23 & 0.24 & 0.51 & 0.19 & 0.35 & 0.35 \\
\bottomrule
\end{tabular}
\end{table}

\begin{table}[h!]
\caption{Mean and standard deviation of sentiment values when the sentiment model does not correctly identifies sentences, but the other model does.}
\label{tab:sentiment_wrong}
\begin{tabular}{llcccccc}
\toprule
\textbf{Label} & & \multicolumn{3}{c}{\textbf{Mean}} & \multicolumn{3}{c}{\textbf{Std}} \\
\cmidrule(lr){3-5} \cmidrule(lr){6-8}
& & \textbf{Positive} & \textbf{Neutral} & \textbf{Negative} & \textbf{Positive} & \textbf{Neutral} & \textbf{Negative} \\
\midrule
\textbf{OBJ} & & 0.23 & 0.37 & 0.39 & 0.14 & 0.32 & 0.40 \\
\textbf{SUBJ} & & 0.29 & 0.37 & 0.32 & 0.23 & 0.36 & 0.34 \\
\bottomrule
\end{tabular}
\end{table}

\subsection{Error Analysis and Language-Specific Challenges}
\begin{figure}[h!]
    \centering
    \includegraphics[width=0.6\linewidth]{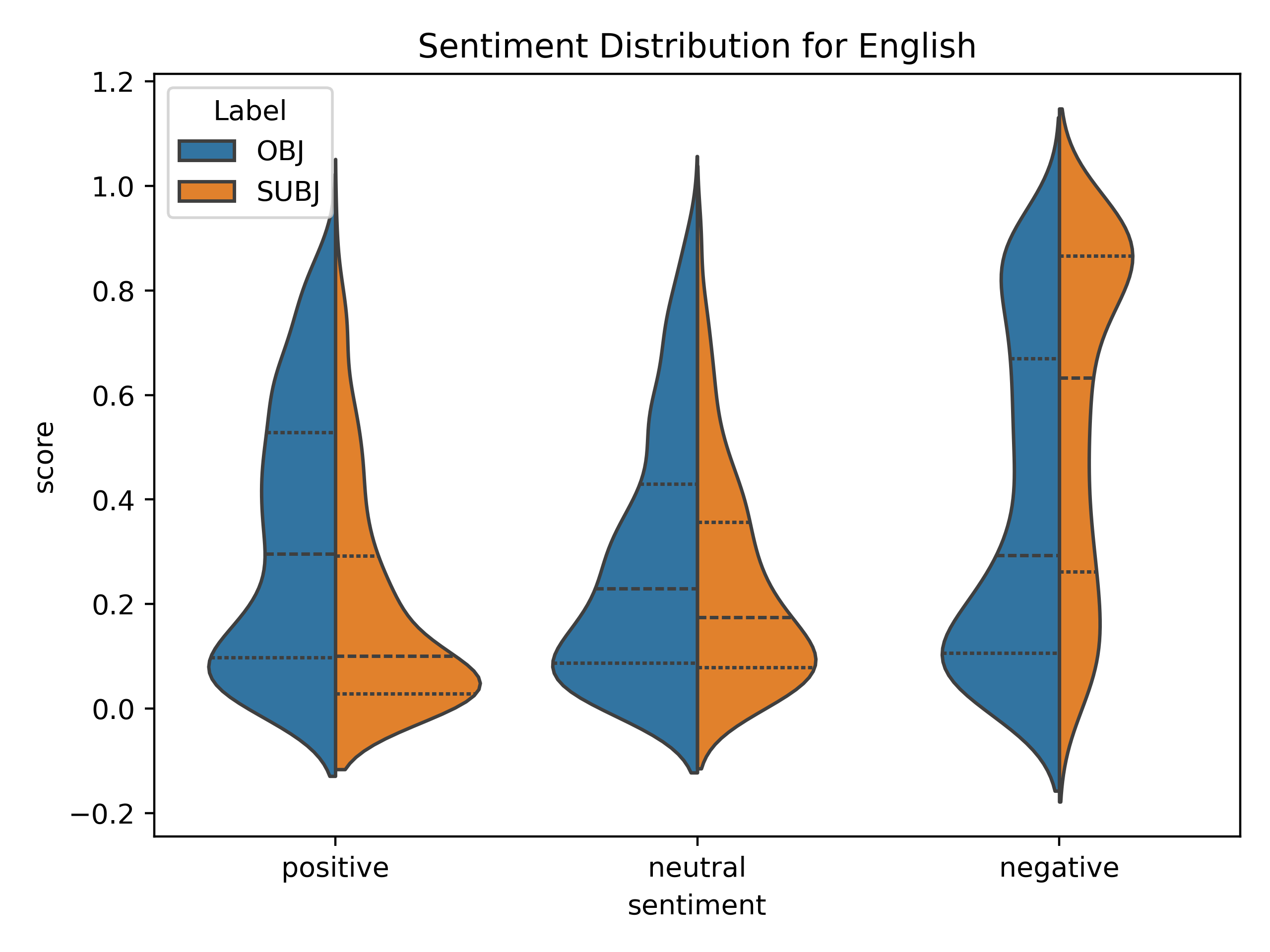}
    \caption{Sentiment distribution over the english language. The three lines in the violin plot represents the first, second and third quartile.)}
    \label{fig:english_distribution}
\end{figure}

\begin{figure}[h!]
    \centering
    \includegraphics[width=0.6\linewidth]{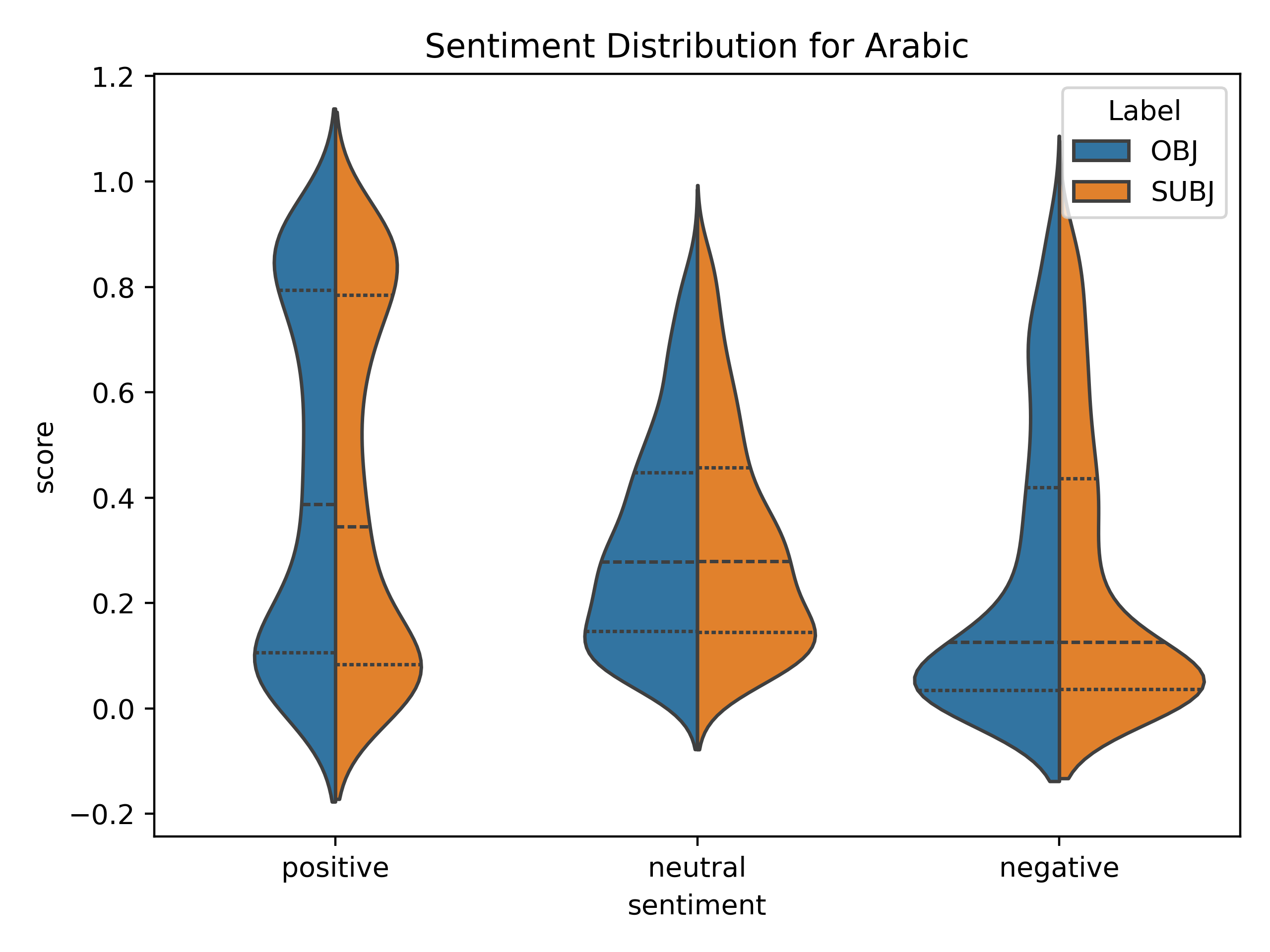}
    \caption{Sentiment distribution over the Arabic language. The three lines in the violin plot represents the first, second and third quartile.)}
    \label{fig:arabic_distribution}
\end{figure}

A consistent challenge across all tasks was the performance on Arabic. Monolingual Arabic models lagged behind others, and including Arabic in multilingual training often diluted overall performance. This suggests that either the pre-trained multilingual embeddings for Arabic are less aligned with this specific task, or that the linguistic expression of subjectivity in the Arabic news sentences provided differs significantly in ways not easily captured by current models without more targeted data or architectural adaptations.
Figure \ref{fig:english_distribution} and Figure \ref{fig:arabic_distribution} (violin plots) illustrate differing sentiment profiles for subjective sentences in English versus Arabic, potentially explaining why sentiment augmentation was more beneficial for some languages than others. More illustrations can be found in Section \ref{sec:appendix}. For English, a high negative sentiment often correlated with subjective labels, a pattern the sentiment-augmented model could leverage. For Arabic, this pattern was less clear or even inverted in the provided dataset, potentially confusing the sentiment-augmented model.
Examples of sentences where sentiment helped:
\begin{itemize}
    \item "But then Trump came to power and sidelined the defense hawks, ushering in a dramatic shift in Republican sentiment toward America’s allies and adversaries." (Sentiment: P:0.109, Ntl:0.035, Neg:0.856) - Strong negative sentiment aided correct SUBJ classification.
    \item "Boxing Day ambush \& flagship attack Putin has long tried to downplay the true losses his army has faced in the Black Sea." (Sentiment: P:0.056, Ntl:0.014, Neg:0.930) - Similarly, high negative sentiment helped.
\end{itemize}

\section{Conclusion}
We presented \textbf{AI Wizards}' system for subjectivity detection in multilingual news articles for the CLEF 2025 CheckThat! Lab Task 1. Our experiments demonstrate that fine-tuned BERT-like architectures, particularly mDeBERTaV3-base, offer robust performance. A key finding is the significant improvement in detecting subjective sentences achieved by augmenting input representations with explicit sentiment scores, especially for languages like English and Italian. Furthermore, decision threshold calibration proved effective for addressing class imbalance, substantially boosting F1 scores on the minority subjective class for languages with skewed distributions.
While explored, Llama3.2-1B in our setup was less competitive than specialized BERT-like models for this task. Performance on Arabic remained a consistent challenge, indicating a need for further research into language-specific modeling or cross-lingual transfer for this language.
Our results highlight the value of combining strong base models with task-relevant feature engineering (sentiment augmentation) and post-processing (threshold calibration) for nuanced NLP problems in multilingual contexts. The code for our system is open-sourced, and a multilingual model incorporating sentiment analysis is available for inference via a Hugging Face dashboard, allowing interactive testing (see Appendix~\ref{sec:appendix} for links). This work contributed to our team achieving high rankings, notably 1st place for Greek (Macro F1~$=~0.51$).

\subsection{Challenge results}
In the following table (Table \ref{tab:challenge_results}), we report our position in all the settings of the challenge that were ranked over a real test set.
\begin{table}[h!]
\centering
\caption{Challenge results - Top 3 scores per category}
\begin{tabular}{|l|l|c|c|}
\hline
\textbf{Setting} & \textbf{Participant} & \textbf{Macro F1} & \textbf{Position} \\
\hline
\multirow{3}{*}{Monolingual - Arabic} 
& aelboua & 0.69 & 1° \\
& tomasbernal01 & 0.59 & 2° \\
& \textbf{AI Wizards} & \textbf{0.56} & \textbf{5°} \\
\hline
\multirow{3}{*}{Monolingual - English} 
& msmadi & 0.81 & 1° \\
& kishan\_g & 0.80 & 2° \\
& \textbf{AI Wizards} & \textbf{0.66} & \textbf{19°} \\
\hline
\multirow{3}{*}{Monolingual - German} 
& smollab & 0.85 & 1° \\
& cepanca\_UNAM & 0.83 & 2° \\
& \textbf{AI Wizards} & \textbf{0.77} & \textbf{5°} \\
\hline
\multirow{3}{*}{Monolingual - Italian} 
& aelboua & 0.69 & 1° \\
& Sumitjais & 0.67 & 2° \\
& \textbf{AI Wizards} & \textbf{0.63} & \textbf{4°} \\
\hline
\multirow{3}{*}{Zeroshot - Greek} 
& \textbf{AI Wizards} & \textbf{0.51} & \textbf{1°} \\
& smollab & 0.49 & 2° \\
& KnowThySelf & 0.49 & 3° \\
\hline
\multirow{3}{*}{Zeroshot - Polish} 
& aelboua & 0.69 & 1° \\
& Sumitjais & 0.67 & 2° \\
& \textbf{AI Wizards} & \textbf{0.63} & \textbf{4°} \\
\hline
\multirow{3}{*}{Zeroshot - Romanian} 
& msmadi & 0.81 & 1° \\
& KnowThySelf & 0.80 & 2° \\
& \textbf{AI Wizards} & \textbf{0.75} & \textbf{7°} \\
\hline
\multirow{3}{*}{Zeroshot - Ukrainian} 
& KnowThySelf & 0.64 & 1° \\
& Ather-Hashmi & 0.64 & 2° \\
& \textbf{AI Wizards} & \textbf{0.64} & \textbf{4°} \\
\hline
\multirow{3}{*}{Multilingual} 
& Bharatdeep\_Hazarika & 0.75 & 1° \\
& kishan\_g & 0.75 & 1° \\
& \textbf{AI Wizards} & \textbf{0.24} & \textbf{15°} \\
\hline
\end{tabular}
\label{tab:challenge_results}
\end{table}

\noindent
Unfortunately, due to an error on our part during the submission process, our multilingual score is very low. As the challenge had already ended, we were unable to correct it. Afterwards, we checked the score we would have achieved, obtaining an Macro F1 score of 0.68: that would have placed us in ninth place.

\section{Limitations}
Our study has several limitations. Sentiment features were derived from a general-purpose model, which may not perfectly capture news-specific subjectivity nuances; its effectiveness also varied by language. The explored Arabic pre-translation introduced potential noise. Computational constraints limited our LLM exploration (Llama3.2-1B); larger or differently fine-tuned LLMs might yield different results. While early fusion of sentiment features during pre-training could offer benefits, our late fusion approach was adopted due to resource constraints. Finally, findings are based on the provided dataset, and generalization to other news sources or subjectivity domains may vary.

\section{Perspectives for Future Work}

Building upon the findings of this work, several promising directions for future research emerge. Our approach highlights the value of sentiment augmentation but also reveals areas for refinement and deeper exploration.

\begin{itemize}
    \item \textbf{Enhanced Sentiment and Emotion Modeling:} The sentiment features used in this study were derived from a general-purpose, Twitter-trained model. Future work could involve fine-tuning a sentiment or emotion analysis model specifically on news corpora to capture more domain-relevant nuances. Exploring more granular emotional features beyond positive/negative/neutral—such as anger, irony, or surprise—could provide even stronger signals for subjectivity. A multi-task learning framework, where a model is simultaneously trained to predict both subjectivity and sentiment/emotion, could also foster a more synergistic learning process.

    \item \textbf{Leveraging Larger Language Models:} Our exploration with Llama3.2-1B was limited by computational constraints. Future research should investigate the capabilities of larger LLMs (e.g., 7B+ parameter models) through more advanced parameter-efficient fine-tuning (PEFT) techniques or full fine-tuning where feasible.
    \item \textbf{Deeper Architectural and Fusion Exploration:} While our simple concatenation (late fusion) of sentiment scores proved effective, more sophisticated fusion mechanisms could yield better performance. Techniques such as attention-based fusion, which would allow the model to dynamically weigh the importance of semantic content versus sentiment signals, warrant investigation. Furthermore, developing interpretability methods to analyze how the model utilizes the concatenated features would provide valuable insights into the decision-making process and help diagnose failures.
\end{itemize}


\section*{Declaration on Generative AI}

\noindent
During the preparation of this work, the author(s) used OpenAI-GPT-4 in order to: grammar and spelling check, paraphrase and reword. After using these tool(s)/service(s), the author(s) reviewed and edited the content as needed and take(s) full responsibility for the publication’s content.

\bibliography{main}

\appendix

\newpage
\section*{Appendix}
\label{sec:appendix}

\subsection*{Dealing with Class Imbalance}
\label{sec:focal-loss}
We also experimented with using \textit{Focal Loss} to address class imbalance in the subjectivity detection task. However, it produced results similar to those obtained using class weights with \textit{Cross-Entropy Loss}, combined with the post-hoc decision threshold calibration employed in our final submissions.

\subsection*{Online Resources}
The source code for our system and models are available at:
\begin{itemize}
    \item GitHub: \href{https://github.com/MatteoFasulo/clef2025-checkthat}{github.com/MatteoFasulo/clef2025-checkthat}
    \item Hugging Face Dashboard (Model Inference): \href{https://huggingface.co/spaces/MatteoFasulo/SubjectivityDetection}{huggingface.co/spaces/MatteoFasulo/SubjectivityDetection}
\end{itemize}

\newpage
\subsection*{Sentiment Distribution}

\begin{figure}[h!]
    \centering
    \includegraphics[width=0.50\linewidth]{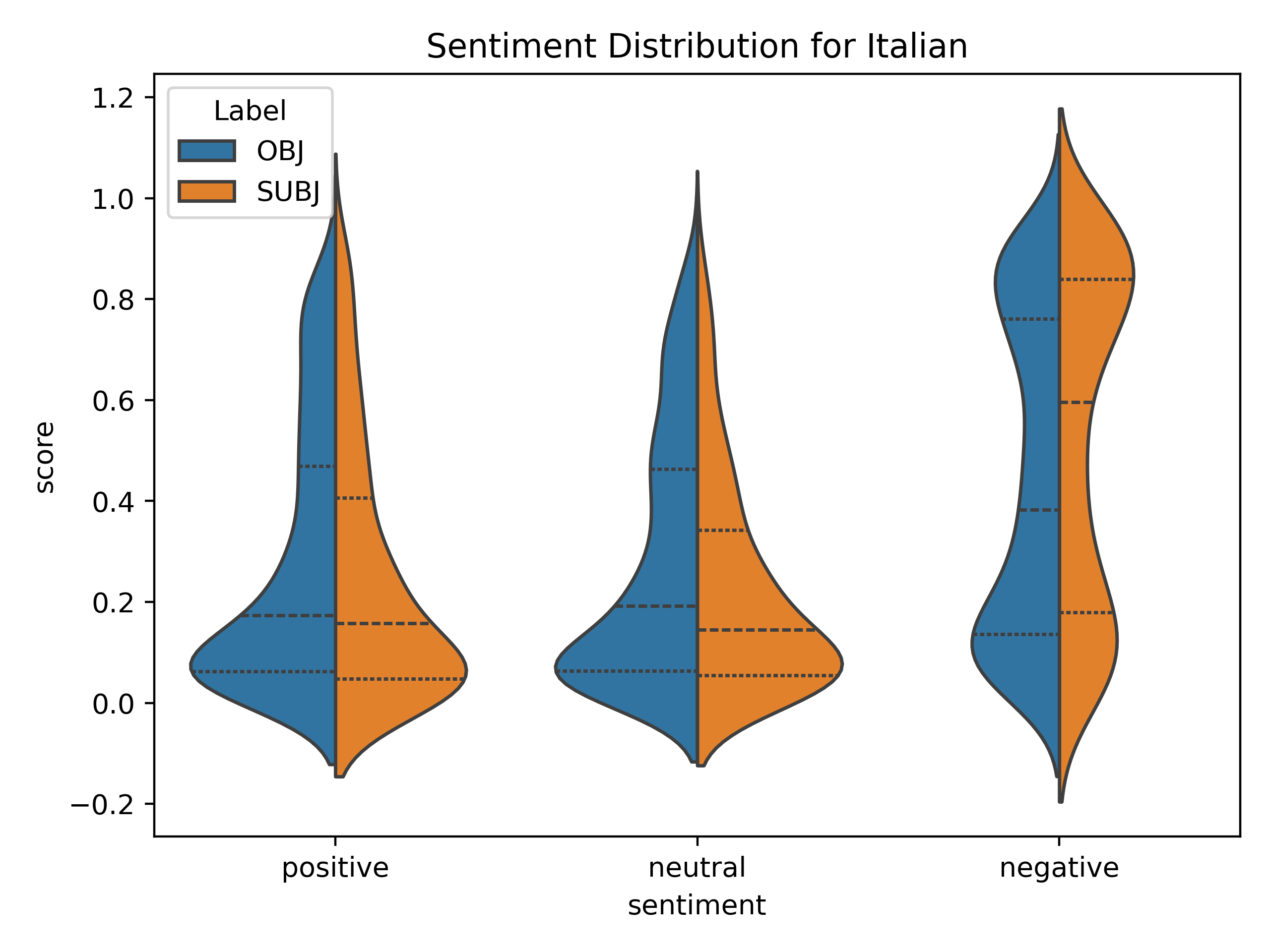}
    \caption{Sentiment distribution over the italian language. The three lines in the violin plot represents the first, second and third quartile.)}
    \label{fig:italian_distribution}
\end{figure}

\begin{figure}[h!]
    \centering
    \includegraphics[width=0.50\linewidth]{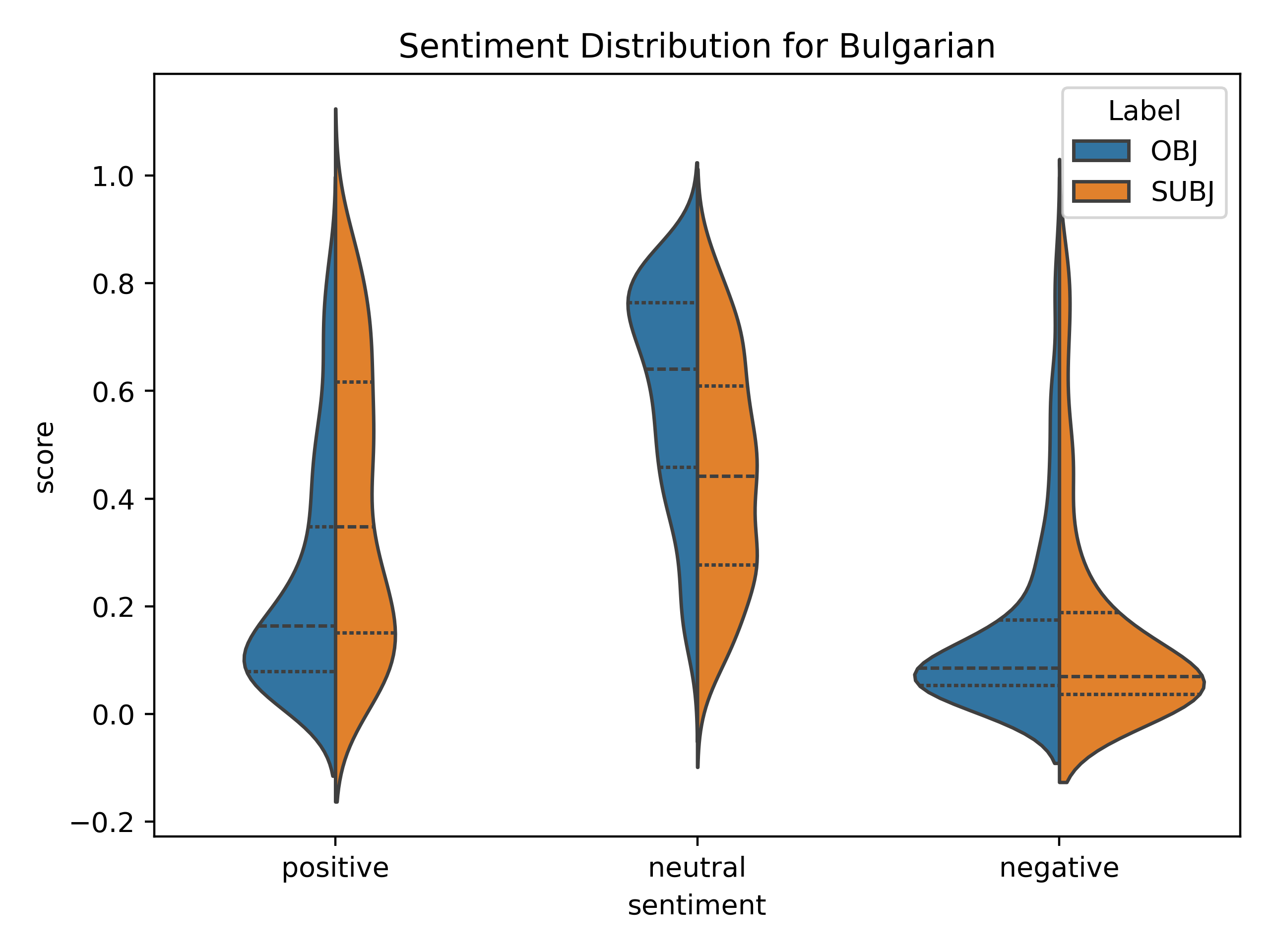}
    \caption{Sentiment distribution over the bulgarian language. The three lines in the violin plot represents the first, second and third quartile.)}
    \label{fig:bulgarian_distribution}
\end{figure}

\begin{figure}[h!]
    \centering
    \includegraphics[width=0.50\linewidth]{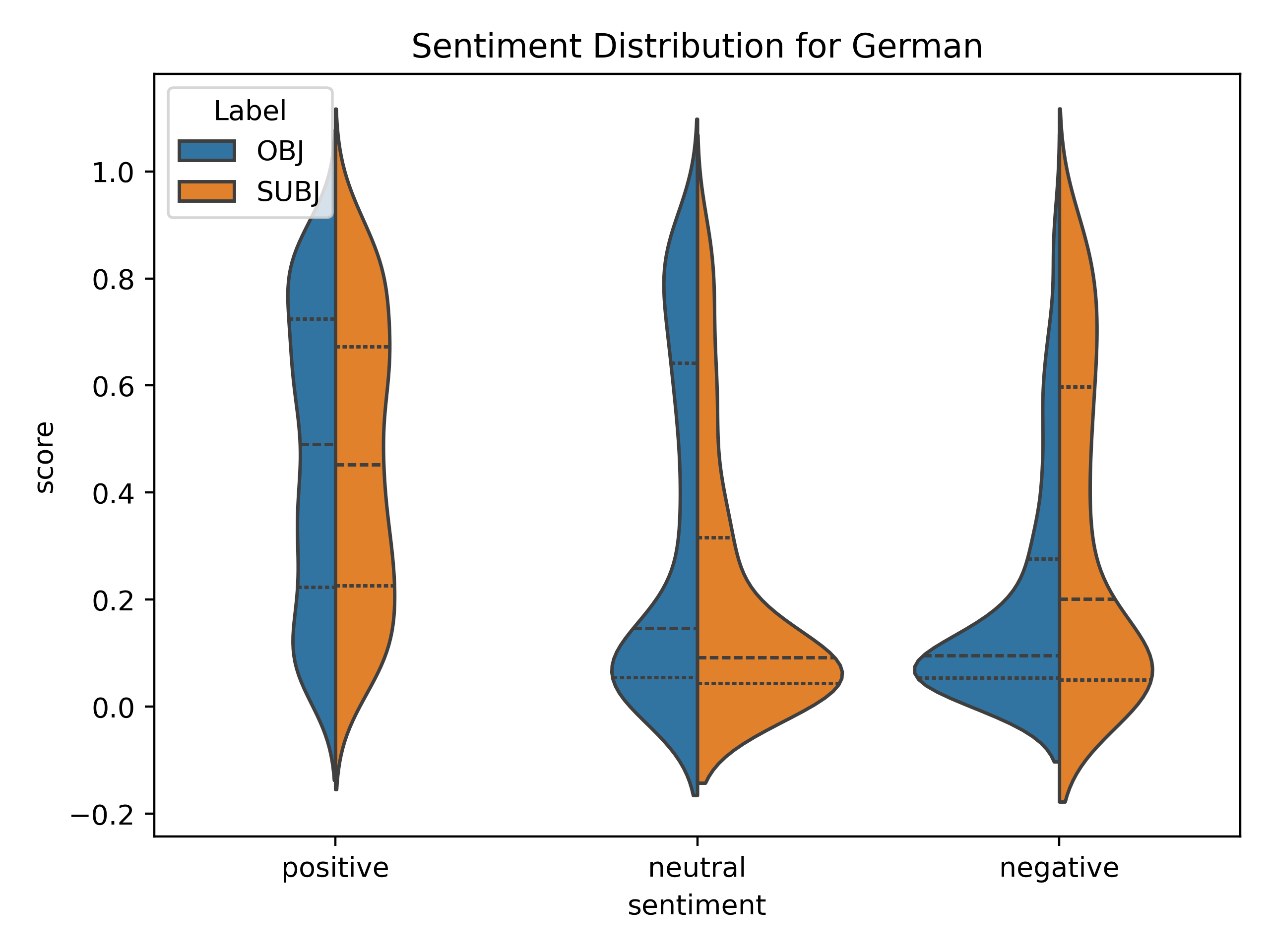}
    \caption{Sentiment distribution over the german language. The three lines in the violin plot represents the first, second and third quartile.)}
    \label{fig:german_distribution}
\end{figure}
\end{document}